# Automated Vision-based Bridge Component Extraction Using Multiscale Convolutional Neural Networks

**Y. Narazaki[1], V. Hoskere[2], T.A. Hoang[3], B.F. Spencer Jr.[4]**

1. *Ph.D. student, Dept of Civil and Environmental Engineering, University of Illinois, Urbana-Champaign, United States.. E-mail: narazak2@illinois.edu*
2. *Ph.D. student, Dept of Civil and Environmental Engineering, University of Illinois, Urbana-Champaign, United States.. E-mail: hoskere2@illinois.edu*
3. *Ph.D. student, Dept of Civil and Environmental Engineering, University of Illinois, Urbana-Champaign, United States.. E-mail: tuhoang2@illinois.edu*
4. *Professor, Dept of Civil and Environmental Engineering, University of Illinois, Urbana-Champaign, United States.. E-mail: bfs@illinois.edu*

**ABSTRACT**
Image data has a great potential of helping post-earthquake visual inspections of civil engineering structures due to the ease of data acquisition and the advantages in capturing visual information. A variety of techniques have been applied to detect damages automatically from a close-up image of a structural component. However, the application of the automatic damage detection methods become increasingly difficult when the image includes multiple components from different structures. To reduce the inaccurate false positive alarms, critical structural components need to be recognized first, and the damage alarms need to be cleaned using the component recognition results. To achieve the goal, this study aims at recognizing and extracting bridge components from images of urban scenes. The bridge component recognition begins with pixel-wise classifications of an image into 10 scene classes. Then, the original image and the scene classification results are combined to classify the image pixels into five component classes. The multi-scale convolutional neural networks (multi-scale CNNs) are used to perform pixel-wise classification, and the classification results are post-processed by averaging within superpixels and smoothing by conditional random fields (CRFs). The performance of the bridge component extraction is tested in terms of accuracy and consistency.

## 1. INTRODUCTION

After earthquakes, bridges need to be inspected rapidly to support the initial response activities, such as life-saving activities, supplying critical goods and services, and providing mass care. Even when the ground motion at the site does not cause structural collapses, rapid inspection and maintenance activities are still important to minimize negative social and economic effects. For example, closure of bridges can cause people unable to go home [1] and refund of train tickets [2]. To reduce the negative effects, safety inspection of the bridges need to be carried out rapidly, followed by the retrofit of any observed damage.

Despite the importance of rapid post-earthquake safety inspection of bridges, earthquakes often cause problems, such as traffic congestion and communication failure [3]. With these problems, human inspectors experience difficulties in accessing the bridges and reporting the safety evaluations they make. To maximize the effect of the post-earthquake bridge safety inspection using limited human resources, the inspection activities need to be planned strategically, based on the initial estimate of the state of the damage to the bridges.

Image data has a potential of assisting the post-earthquake bridge safety inspection by automating the initial damage assessment. Image data can be acquired easily and quickly using consumer digital cameras, or potentially using more advanced systems, such as satellite imagery [4] and unmanned aerial vehicles (UAVs) [5] [6]. Besides, image data captures the visual information, which is also evaluated during the post-earthquake visual inspection by human inspectors. Therefore, by establishing and implementing appropriate steps of data processing, image data is expected to provide initial knowledge of the current state of the structures without interaction with the human experts.

The applications of vision-based techniques to the civil engineering structures can be classified into two levels, i.e. component level and entire structure level. In the component level, damage detection problems such as crack detection [7] [8] [9] and spalling detection [10] have been investigated using local contexts encoded by different methods. However, those damage detection approaches assume a close-up image of a single component, and distinguish damaged surfaces from healthy surfaces. Therefore, those methods are not guaranteed to work for images containing multiple components of different structures, often raising significant number of false-positive alarms [8]. Compared to the component-level approaches, research works about damage evaluation of entire structures are relatively immature. Although convolutional neural networks (CNN) have been applied to assess the structural conditions of buildings after earthquakes [11], the approach is currently limited to the binary evaluation of whether the buildings are damaged or not. The evaluation of component-level damage from images of entire structures in complex scenes has not been achieved because of the complexity of the problem. Damage evaluation based on three-dimensional information is not reviewed here because of the computational intensity of data processing, making the approach less effective for post-earthquake inspections.

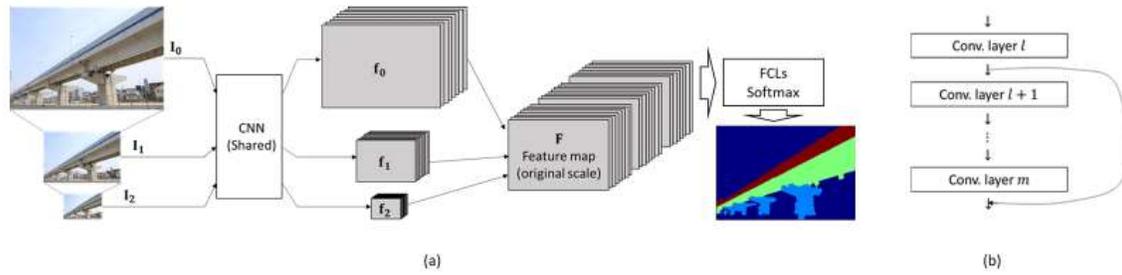

Figure 2.1 Pixel-wise labelling techniques (a)Multi-scale CNN [10] (b)Residual networks [11]

To fill the gap between images of complex scenes and the component-level vision-based damage evaluation methods, an approach for the automatic detection and localization of critical structural components need to be developed. Previously, an image-based concrete column recognition method based on line segment detection has been proposed [12]. However, the method does not encode higher-level contextual information, which is necessary to recognize structural components from images of complex scenes. More recent study [13] uses multi-scale convolutional neural networks (multi-scale CNNs) to recognize bridge components using high-level scene understanding. Although this approach is shown to be effective, the raw output from the multi-scale CNNs contains isolated or unsmooth component detections. To refine the detection results, this study implements post-processing methods on top of the bridge component recognition approach using multi-scale CNNs. This paper first describes the methods implemented to recognize bridge components, i.e. multi-scale CNNs [14], SLIC superpixel algorithm [15] [16], and conditional random field (CRF) modeling and optimization [14] [17] [18], including the implementation details. Then, the training procedures and the testing results of the multi-scale CNNs are described. Finally, the post-processing methods are applied and the effect of post-processing is discussed.

## 2. VISION-BASED BRIDGE COMPONENT EXTRACTION APPROACH

In this study, bridge components in images of complex scenes are extracted using three-step approach presented by Farabet et al. [14]. In this approach, every pixel in an image is first classified into five bridge component classes using multi-scale convolutional neural networks (multi-scale CNNs). Then, the classification results of the multi-scale CNNs are post-processed by averaging the results within small segments called superpixels. Finally, the pixel labels are further refined by optimizing the conditional random field (CRF) model. This section describes the details of each step.

### 2.1. Multi-scale convolutional neural networks for pixel-wise labeling

In this study, bridge components are extracted based on the pixel-wise classification using the multi-scale convolutional neural networks (Multi-scale CNNs). Multi-scale CNNs [14] was designed to recognize complex scene objects photographed in different scales. As shown in Fig. 2.1(a), the multi-scale CNNs achieves approximate scale invariance for the feature extraction by applying convolutional and pooling layers shared among images downsampled to multiple resolutions. After extracting the feature maps in multiple resolutions, the feature maps are upsampled to the original resolution, and concatenated to form a multi-scale dense feature map, **F**. The dense feature map **F** can be represented by a three-dimensional matrix, whose first and second dimensions indicate the pixel location within image, and the last dimension indicates features at the specified pixel location. Then, the feature vector at each pixel location is classified into an appropriate category by applying fully-connected layers (FCLs), a softmax layer, and the argmax of the softmax probabilities.

This method was previously applied to the automated bridge column recognition task [13]. In the study, four different network architectures ranging from 5 to 45 layers are compared using a combined 10-class scene classification dataset [19] [20] [21] [22] [23] [24] [25]. Then, the network with a 22-layer shared CNN and two FCLs are selected based on the test accuracy. Following the study, this study uses the similar network architecture with the 22-layer shared CNN and one FCL (Table 2.1). Each convolutional layer consists of a convolution using the specified size of filters, followed by the ReLU activation function and batch normalization [26]. Besides, the network has the architecture of a residual network (ResNet) [27], which is characterized by adding previous layer output(s) to the current layer output (Fig. 2.1(b)). The ResNet connections are known to be effective when the network has a deep architecture. Compared to the architecture used in the previous work [13], the number of FCLs is reduced to one to save the network size and take advantage of deep architecture.

To perform pixel-wise bridge component recognition on a complex scene image, the two-step classification approach [13] is implemented. In this approach, the classification begins with applying a multi-scale CNN to compute the softmax probabilities corresponding to 10 scene classes (Building, Greenery, Person, Pavement, Sign & Poles, Vehicles, Bridges, Water, Sky, and Others). Then, the softmax probability maps of the first nine classes are scaled to [0,255] and concatenated with the raw RGB image to form an augmented input to the component classifier. The component classifier estimates the softmax probabilities and the final pixel labels using another multi-scale CNN. This two-step approach is effective in reducing false-positives when the image contains multiple scene objects.

## 2.2 Post-processing Steps

Although the multi-scale CNN is effective in recognizing bridge components within complex scene images, the raw outputs from the multi-scale CNN still contains inconsistent detections, such as isolated false positive detections and the structural members which are not continuous with any other component. To enhance the accuracy and consistency of the bridge component recognition results, post-processing methods used by Farabet et al. [14] are applied to the multi-scale CNN output.

The post-processing contains two steps, superpixel averaging and conditional random field (CRF) optimization. In the first step, the softmax probabilities computed by the multi-scale CNN are averaged over small segments called superpixels. Then, the segment labels are optimized using the CRF model. The details of each step are described below.

### 2.2.1 Step 1: Superpixel averaging

In this step, the softmax probabilities computed by the multi-scale CNN are averaged over small segments called superpixels. In this study, the superpixels are generated by SLIC algorithm [16] [15], which clusters the pixels in CIELAB color space using the distance measure expressed by

$$D = \sqrt{d_c^2 + \left(\frac{d_s}{S}\right)^2 m^2} \qquad (2.1)$$

In this expression, $d_c$ and $d_s$ are the color proximity and spatial proximity. The parameter $S$ in Eq. 2.1

Table 2.1 ResNet22 architecture

| ResNet22 architecture | | | | | |
|---|---|---|---|---|---|
| Name | Filt. Size | ResNet connect. | Name | Filt. Size | ResNet connect. |
| Conv0 | 7x7x64 | | Conv10 | 3x3x128 | Maxpool1 |
| Maxpool0 | 2x2 | | Conv11 | 3x3x128 | |
| Conv1 | 3x3x64 | | Conv12 | 3x3x128 | Conv10 |
| Conv2 | 3x3x64 | Maxpool0 | Conv13 | 3x3x128 | |
| Conv3 | 3x3x64 | | Conv14 | 3x3x128 | Conv12 |
| Conv4 | 3x3x64 | Conv2 | Conv15 | 3x3x128 | |
| Conv5 | 3x3x64 | | Conv16 | 3x3x128 | Conv14 |
| Conv6 | 3x3x64 | Conv4 | Conv17 | 3x3x128 | |
| Conv7 | 3x3x64 | | Conv18 | 3x3x128 | Conv16 |
| Conv8 | 3x3x64 | Conv6 | Conv19 | 3x3x128 | |
| Maxpool1 | 2x2 | | Conv20 | 3x3x128 | Conv18 |
| Conv9 | 3x3x128 | | FCL0 | 10 or 5 | |
| Batch size | 10 ||||||
| Wt. decay | 0.0001 ||||||
| Dropout | None ||||||
| #param | 2003904 (scene), 2030208 (component classifier with scene) ||||||

indicates the approximate distance between cluster centers, and $m$ is a compactness parameter. The SLIC method first initialize the specified number of cluster centers uniformly, and iteratively update the cluster centers using the distance measure $D$ from the cluster centers. Compared to other superpixel algorithms, this method can provide low-level segmentations in a computationally efficient manner [16].

The softmax probability output of the multi-scale CNN at pixel $i$ and label $l$, $\hat{d}_{i,l}$, is averaged using the superpixels computed by the SLIC algorithm. For example, the softmax probabilities in segment $S_j$ are averaged by

$$\hat{d}_{sp_{j,l}} = \frac{1}{|S_j|} \sum_{i \in S_j} \hat{d}_{i,l} \qquad (2.2)$$

where $|S_j|$ is the number of pixels in the segment $S_j$, and $\hat{d}_{sp_{j,l}}$ is the averaged probability for the segment $j$ and label $l$. This step has an effect of removing the change of the assigned labels below the scale of the superpixels.

### 2.2.2 Step 2: Conditional Random Field Optimization

The first post-processing step removes the change of assigned labels within the superpixels to reduce the inconsistent labels. However, the label refinement by the superpixel averaging is limited to the inconsistent labels below the scale of superpixels. To impose consistency in larger scales, conditional random field (CRF) model is applied to the results of the superpixel averaging step [14].

CRF model for the post-processing is formulated by the energy function with unary and binary terms as follows.

$$E = \sum_{i \in segments} U_i(L_i) + \sum_{\{i,j\} \in Neighboring\ segment\ pairs} B_{ij}(L_i, L_j) \qquad (2.3)$$

where $L_i$ and $L_j$ are the labels assigned to the segment $i$ and $j$. The refined segment labels are determined by approximately minimizing the total energy function $E$ within the image.

In this study, the unary and the binary terms are defined as

$$U_i(L_i) = \sum_{p \in S_i} \exp\left(-\alpha \hat{d}_{p,L_p}\right) [L_p \neq L_{sp_i}]$$

$$B_{ij}(L_i, L_j) = \sum_{\substack{p \in S_i, q \in S_j \\ \{p,q\} \in Neighboring\ pixel\ pairs}} \gamma w_{L_i L_j} \exp\{-\beta(\|\nabla I\|_p + \|\nabla I\|_q)/2\}[L_i \neq L_j] \qquad (2.4)$$

where $p$ and $q$ indicate pixels, $L_{sp_i}$ is the label of the $i^{th}$ segment assigned by the superpixel averaging, $[\cdot]$ is an indicator function, which takes the value of 1 when the statement inside the bracket is true, and 0 otherwise. The $\alpha, \beta$, and $\gamma$ are the parameters, and $\|\nabla I\|_p, \|\nabla I\|_q$ are the gradient magnitude at pixel $p$ and $q$, respectively. Different from Farabet et al. [14], the binary terms have an additional weight $w_{L_i L_j}$, which takes values of 0.5 or 1, depending on the combination of labels between the neighboring segments. The energy function thus derived is minimized using the $\alpha$-expansion algorithm [18] [17].

## 3. BRIDGE COMPONENT RECOGNITION RESULTS

Based on the implementation described in the previous section, the multi-scale CNNs are trained and tested, followed by the two-step post-processing using superpixel averaging and CRF optimization. In this study, the multi-scale CNN operations and the post-processing methods are implemented using Tensorflow [28] and Python. This section describes the setting and the results of the bridge component recognition.

### 3.1. Datasets

The scene and bridge component datasets described by Narazaki et al. [13] are used to train and test the multi-scale CNNs for the bridge component recognition. The scene classification dataset consists of 3403 general images, 6842 urban images, and 1652 bridge images. These images are collected by combining existing fully-labeled datasets (Stanford Background Dataset [23], SIFT Flow Dataset [24], SYNTHIA Dataset [25], CamVid Dataset [20] [19]) and images labeled by the authors' research group. For the general and urban categories, 90% of the images are used for training and the remaining 10% of the images are used for testing. For the bridge category, 234 testing images are selected such that the training and the testing images are independent. The combined scene classification dataset has 10 classes, i.e. Building, Greenery, Person, Pavement, Sign & Poles, Vehicles, Bridges, Water, Sky, and Others. The general and urban images are fully labeled into the 10 classes by manual process, or by appropriate transfer of existing labels. In contrast, the bridge images without existing labels are labeled only partly, some of which are labeled only in the bridge part. The bridge component classification dataset consists of 1329 training images and 234 testing images from the same sources. In this dataset, the images are fully labeled into 5 classes, i.e. Non-Bridge, Columns, Beams & Slabs, Other structural, and Other nonstructural. The details of the datasets are explained in [13].

### 3.2 Scene Classification Results

A multi-scale CNN with *ResNet22* architecture shown in Table 2.1 is trained first to provide high-level scene understanding to the component classifier. The network parameters are updated by Adam algorithm [30] using mini-batches. The mini-batches are made by randomly selecting four images from general category, four images from urban category, and two images from bridge category. To improve the training results, batch normalization [26], weight decay of 0.0001 [31], and median frequency balancing [32] are implemented. Besides, the image data are augmented by random resizing (between 75% and 125% for general and the urban images, between 50% and 150% for bridge images), random cropping (180×180 images are cropped from 320×320 images), random rotation up to ±15°, random flipping, and jitter (zero-mean normal distribution with $\sigma = 2$).

During the training, the number of iterations are counted by cycles, which refer to approximately 1.4k iterations until the network sees all images in the dataset. The multi-scale CNN is trained for 100 cycles in total with gradually decreasing learning rate. For the first 50 cycles, the learning rate was set to 0.001, followed by 35 cycles of iteration with learning rate 0.0001 and 15 cycles of iterations with learning rate 0.00001.

The confusion matrix of the trained multi-scale CNN is shown in Fig. 3.1(a). The accuracies for the eight out of 10 scene classes exceed 80%, including over 90% accuracy for bridges. The total pixel-wise accuracy is 88.27%. The softmax probability maps computed by this scene classifier is used with the original images to train the bridge component classifier.

### 3.3 Bridge Component Classification Results

On top of the pretrained scene classifier, another multi-scale CNN with *ResNet22* architecture is trained using the same techniques and data augmentation methods. The number of iterations are counted in terms of cycles, which refer to approximately 130 iterations until the network sees all images in the bridge dataset. By using mini-batches of 10 randomly selected images, the network weights are updated for 1000 cycles in total. The learning rate during training is set to 0.001 for the first 500 cycles, 0.0001 for the next 400 cycles, and 0.00001 for the last 100 cycles. The confusion matrix of the bridge component classifier with scene understanding is shown in Fig. 3.1(b). The total pixel-wise accuracy for testing images is 78.95%. The accuracy is comparable with the testing accuracy in the previous work [15], even without the hidden FCL after the convolutional layers.

The bridge component classifier with scene classification is also applied to non-bridge test images in the scene classification dataset to evaluate the false-positive detections. For comparison, a *ResNet22* network without scene understanding is also trained in the same manner and applied to the same non-bridge images (Naïve approach in [15]). The rates of the false-positive bridge component detections are shown in Table 3.3. The table shows significantly less false-positives when the bridge component recognition is combined with the scene classification results. Some of the example recognition results are visualized in Fig. 3.2.

### 3.4 Superpixel Averaging

As a first step of post-processing the raw multi-scale CNN output, the softmax probabilities are averaged within the superpixels generated by the SLIC algorithm discussed previously. After observing the superpixels generated with some parameter values, the compactness factor $m$, the approximate number of segments, and the number of iterations are set to 5, 500, and 50, respectively.

Example superpixel averaging results are shown in Fig. 3.3(a). For the upper right and the lower right images, small isolated false positive detections are removed by this processing. However, errors in the superpixel segmentation have significant effects on the results after superpixel averaging. For example, the boundary of the cables and piers of the cable-stayed bridge is less accurate after the post-processing.

### 3.5 Optimization of Conditional Random Field Model

In this step, the labels are refined by minimizing the CRF model discussed previously. Before the energy minimization, the parameters of the CRF model are tuned using 10% of the testing images in the bridge component classification dataset. By evaluating the accuracy for different sets of the parameter values including all possible combinations of 0.5 and 1.0 for the weight $w_{L_iL_j}$, the parameters $\alpha, \beta$, and $\gamma$ are set to 0.1, 20.0, 10.0, respectively. The weight values for each combination of labels, $w_{L_iL_j}$, are determined as shown in Table 3.2.

The class accuracies are evaluated using 90% of the testing images of the bridge component classification dataset

Table 3.1 False-positive rates for the 9 non-bridge classes with and without scene classification

| Scene label | Bldg | Green | Person | Pvmt | S&P | Vhcls | Water | Sky | Other |
|---|---|---|---|---|---|---|---|---|---|
| FP (w/ scene) [%] | 1.8 | 1.1 | 10.8 | 5.2 | 3.9 | 3.8 | 1.9 | 0.2 | 3.3 |
| FP (w/o scene) [%] | 53.0 | 20.6 | 39.1 | 30.1 | 38.9 | 30.4 | 23.1 | 3.9 | 40.5 |

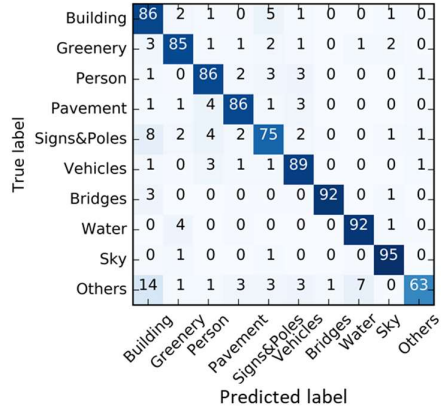
(a)

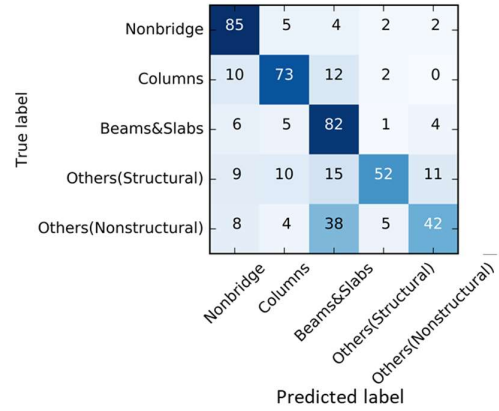
(b)

Figure 3.1 Confusion matrices (a)Scene classifier (b)Bridge component classifier

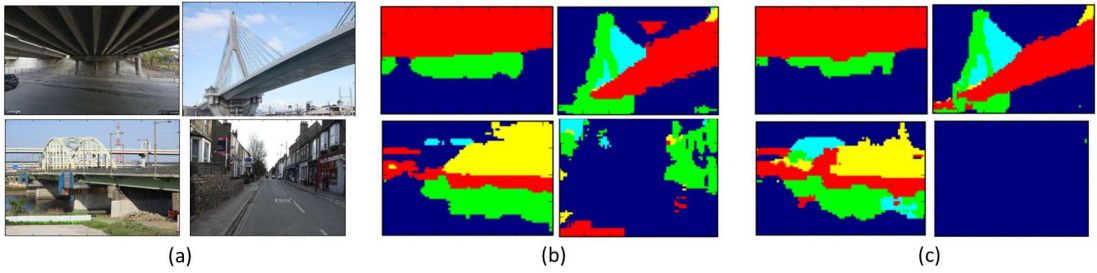

(a)      (b)      (c)

Figure 3.2 Example bridge component recognition results by multi-scale CNNs
(a)Input images (b)Results without scene classifier (c)Results with scene classifier

Table 3.2, Weights $w_{L_i L_j}$ in the binary term of the CRF energy function

| $L_i$ \ $L_j$ | Nonbridge | Columns | Beams&Slabs | Other structural | Other nonstructural |
|---|---|---|---|---|---|
| Nonbridge | 0.0 | 0.5 | 1.0 | 1.0 | 0.5 |
| Columns | 0.5 | 0.0 | 1.0 | 0.5 | 1.0 |
| Beams&Slabs | 1.0 | 1.0 | 0.0 | 0.5 | 0.5 |
| Other structural | 1.0 | 0.5 | 0.5 | 0.0 | 1.0 |
| Other nonstructural | 0.5 | 1.0 | 0.5 | 1.0 | 0.0 |

not used in the CRF parameter tuning. As shown in the example results in Fig. 3.3(b), the CRF optimization enhances the consistency of the bridge component recognition by removing some of the isolated or unsmooth labels. However, the CRF optimization also removes a part of the pier of the cable-stayed bridge. Furthermore, the errors in the superpixel segmentation have a significant effect on the CRF optimization as well. The total pixel-wise testing accuracy is 78.94%, which is about 0.01% lower than the raw multi-scale CNN results. The confusion matrix of the bridge component recognition after CRF optimization is shown in Fig. 3.3(c).

From the results presented in this section, the two-step bridge component recognition using multi-scale CNNs, followed by the post-processing using superpixels and CRF models, are shown to be effective in detecting, localizing, and extracting bridge components from complex scene images. Integrating the scene information has an effect of reducing the false-positive detections. Superpixel averaging of the raw multi-scale CNN output refines the classification results by smoothing the assigned labels in the scale of the superpixels. Finally, the CRF optimization provides smoothness of labeling in larger scales by minimizing the total energy within an image. The post-processing steps applied in this study is shown to be effective in removing inconsistent labels, such as

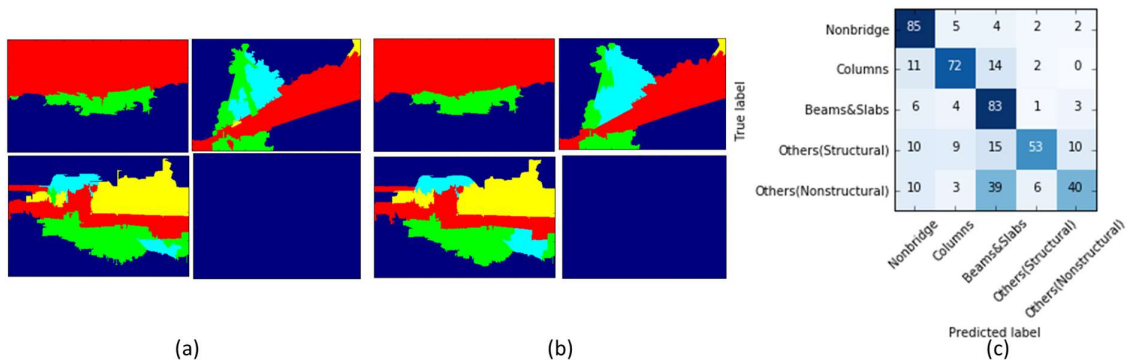

Figure 3.3 Example post-processing results
(a)Superpixels averaging examples (b)CRF optimization examples (c)Confusion matrix after CRF optimization isolated and unsmooth labels. However, the errors in the superpixel segmentation have a significant effect on the recognition results. Therefore, post-processing methods which is not directly dependent on the superpixel segmentation need to be investigated in the future.

## 4. CONCLUSIONS

This study showed an approach to the automated vision-based bridge component extraction to provide the component-level automatic damage detection methods with the areas to focus on within complex scene images. The bridge component extraction starts by computing the softmax probability maps corresponding to the 10 scene classes using a multi-scale CNN with 22 layers. Then, the input image is augmented by the softmax probability maps and fed into another multi-scale CNN to classify every pixel into one of the five component classes. Finally, the estimated labels are refined by averaging the softmax probabilities over superpixels and optimizing the labels such that the CRF energy function is minimized.

The pixel-wise classification integrated with the scene classification was shown to have an effect of reducing false positive bridge component detections in complex scene images. The superpixel averaging and the CRF model optimization were also shown to be effective in removing isolated or unsmooth labels. The total pixel-wise accuracy after post-processing was 78.94%.

A problem of the implemented approach is that the recognition results are strongly dependent on the superpixel segmentation results. When the superpixel does not follow the boundary of objects in the scene, the resulting recognition results are erroneous. To address this problem, an approach which does not directly depend on the superpixel segmentation need to be investingated in the future.